\documentclass[letterpaper]{article} 
\usepackage{aaai23}  
\usepackage{times}  
\usepackage{helvet}  
\usepackage{courier}  
\usepackage[hyphens]{url}  
\usepackage{graphicx} 
\usepackage{natbib}  
\usepackage{caption} 
\usepackage{algorithm}
\usepackage{algorithmic}

\usepackage{newfloat}
\usepackage{listings}
\usepackage{amssymb, bm, amsmath}
\usepackage{multirow}
\usepackage{subfig,}
\usepackage{xcolor}
\usepackage{eucal}
\usepackage[switch]{lineno}
\DeclareCaptionStyle{ruled}{labelfont=normalfont,labelsep=colon,strut=off} 
\floatstyle{ruled}
\newfloat{listing}{tb}{lst}{}
\floatname{listing}{Listing}
\title{DINet: Deformation Inpainting Network for Realistic Face Visually Dubbing on High Resolution Video}
\author{
    Zhimeng Zhang\textsuperscript{\rm 1}\footnotemark[1],
    Zhipeng Hu\textsuperscript{\rm 1 3}\footnotemark[1],
    Wenjin Deng\textsuperscript{\rm 2}\footnotemark[1],
    Changjie Fan\textsuperscript{\rm 1},
    Tangjie Lv\textsuperscript{\rm 1},
    Yu Ding\textsuperscript{\rm 1 3}\thanks{Equal contribution. Yu Ding is the corresponding author.} \\
}
\affiliations{
    \textsuperscript{\rm 1} Virtual Human Group, Netease Fuxi AI Lab \\
    \textsuperscript{\rm 2} Xiamen University \\
    \textsuperscript{\rm 3} Zhejiang University \\
    \{zhangzhimeng, zphu, fanchangjie, hzlvtangjie, dingyu01\}@corp.netease.com \\
    dengwenjin@stu.xmu.edu.cn 
}

\usepackage{bibentry}

\begin{document}

\maketitle

\begin{abstract}
For few-shot learning, it is still a critical challenge to realize photo-realistic face visually dubbing on high-resolution videos. Previous works fail to generate high-fidelity dubbing results. To address the above problem, this paper proposes a Deformation Inpainting Network (DINet) for high-resolution face visually dubbing. Different from previous works relying on multiple up-sample layers to directly generate pixels from latent embeddings, DINet performs spatial deformation on feature maps of reference images to better preserve high-frequency textural details. Specifically, DINet consists of one deformation part and one inpainting part. In the first part, five reference facial images adaptively perform spatial deformation to create deformed feature maps encoding mouth shapes at each frame, in order to align with the input driving audio and also the head poses of the input source images. In the second part, to produce face visually dubbing, a feature decoder is responsible for adaptively incorporating mouth movements from the deformed feature maps and other attributes (i.e., head pose and upper facial expression) from the source feature maps together. Finally, DINet achieves face visually dubbing with rich textural details. We conduct qualitative and quantitative comparisons to validate our DINet on high-resolution videos. The experimental results show that our method outperforms state-of-the-art works. 
\end{abstract}

\section{Introduction}

Talking head generation tasks, including one-shot talking face \cite{chung2017you,chen2018lip,zhou2019talking,vougioukas2020realistic,chen2019hierarchical,song2018talking,das2020speech,chen2020talking,zhou2020makelttalk,zhou2021pose,zhang2021flow,wang2021audio2head,wang2022one}, person-specific talking face \cite{lahiri2021lipsync3d,guo2021ad,zhang2021facial,wu2021imitating,fried2019text,song2022everybody,thies2020neural,ji2021audio} and few-shot face visually dubbing \cite{kr2019towards,prajwal2020lip,xie2021towards,park2022synctalkface}, have attracted growing research attention due to broad applications in media production, film industry, etc. 

\begin{figure}
\centering
\includegraphics[width=0.45\textwidth]{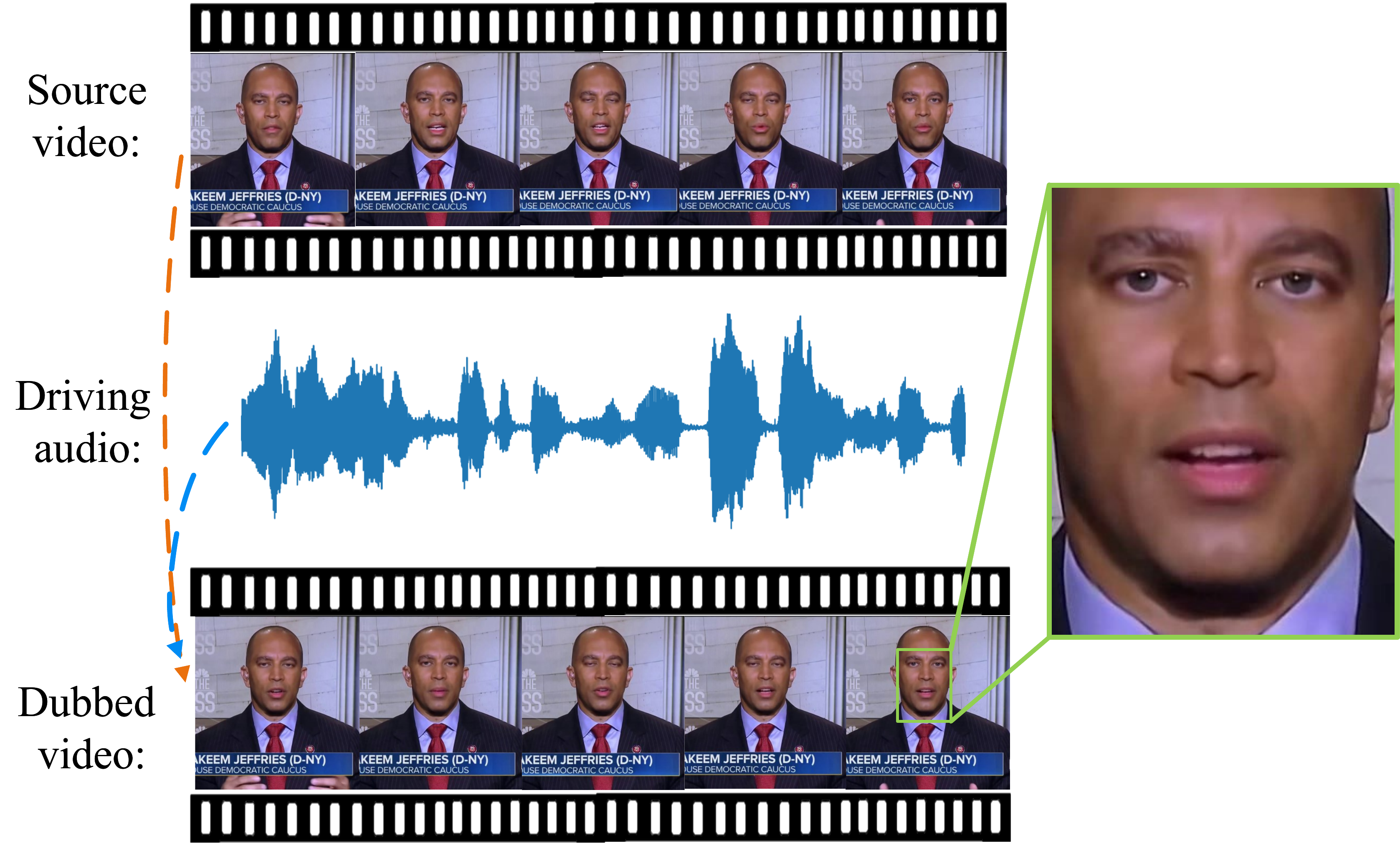}
\caption{Our method dubs a high-resolution source video according to a driving audio.}
\label{fig:teaser}
\end{figure}

Face visually dubbing, as shown in Figure \ref{fig:teaser}, aims to synchronize mouth shape in a source video according to an input driving audio, while keeping identity and head pose in consistent with the source video frame. Existing few-shot face visually dubbing works \cite{kr2019towards,prajwal2020lip,xie2021towards,park2022synctalkface,liu2022parallel} make great efforts on synthesizing realistic faces. They utilize convolutional networks with multiple up-sample layers to directly generate mouth pixels from latent embeddings. For example, Wav2Lip \cite{prajwal2020lip} designs face decoder with six de-convolutional layers. TRVD \cite{xie2021towards} combines up-sample layers with AdaIN \cite{huang2017arbitrary} in face decoder. However, these straightforward methods always face trouble on producing high-resolution videos, despite using high-resolution training data (see the results of Wav2Lip \cite{prajwal2020lip} in Figure~\ref{fig:qua_comp})

One main reason is that, under a few-shot condition, mouth textural details have few correlations with driving audios, making it challenging to generate high-frequency textural details directly. For networks, it is difficult to learn a complex mapping of audio-lip synchronization from amount of identities with various speaking styles, thus they easily ignore correlated textural details. Furthermore, compared with person-specific dubbing works \cite{fried2019text,thies2020neural}, this reason makes few-shot works always generate more obvious blur results, when the person-specific dubbing works also rely on straightforward generation methods.

To solve the above problem, we develop spatial deformation on feature maps of reference facial images to inpaint mouth pixels. Specifically, our spatial deformation is able to synchronize a mouth shape with a driving audio and align a head pose with a source face. Deformation operation moves pixels into appropriate locations rather than generation from scratch, thus it nearly preserves all textural details. 

In this paper, we propose a Deformation Inpainting Network (DINet) for realistic face visually dubbing on high-resolution videos. The framework is shown in Figure \ref{fig:pipeline}. DINet consists of two parts: a deformation part and an inpainting part. The deformation part first encodes the features of head pose and speech content from the source face and the driving audio respectively, and then utilizes these features to deform the reference faces. The inpainting part merges features of source face and deformed results by convolutional layers to inpaint the pixels in source mouth region. With the combination of deformation and inpainting, our DINet achieves more realistic face visually dubbing than direct-generation based methods.

Our contributions are summarized as follows:
\begin{itemize}

\item We develop and validate a novel Deformation Inpainting Network (DINet) to achieve face visually dubbing on high-resolution videos. Our DINet is able to produce accurate mouth movements but also preserve textual details. 
    \item We conduct qualitative and quantitative experiments to evaluate our DINet, and experimental results show that our method outperforms state-of-the-art works on high-resolution face visually dubbing.  
    
\end{itemize}

\section{Related Work}

\subsection{Talking Face Generation}
Talking face generation aims to synthesize facial images according to a driving audio or text. It consists of three main directions: one-shot talking face, few-shot face visually dubbing and person-specific talking face.

\textbf{One-shot talking face.} One-shot talking face focus on driving one reference facial image with synchronic lip movements, realistic facial expressions and rhythmic head motions. Some works utilize latent embeddings to generate talking face. They first encode a reference image and a driving audio into latent embeddings, and then use networks to decode the embeddings into a synthetic image. Extra losses, like deblurring loss \cite{chung2017you}, audio-visual correlation loss \cite{chen2018lip}, audio-visual disentangled loss \cite{zhou2019talking,zhou2021pose,liang2022expressive} and spatial-temporal adversarial loss \cite{song2018talking,vougioukas2020realistic} are used to improve lip synchronization and visual quality.

Other works leverage explicit intermediate representations, including unsupervised keypoints \cite{wang2021audio2head,wang2022one,ji2022eamm}, facial landmarks \cite{chen2019hierarchical,das2020speech,zhou2020makelttalk} and 3DMM \cite{chen2020talking,zhang2021flow} to synthesize facial images. They split a pipeline into two separated parts: one audio-to-animation part and one animation-to-video part. Two parts are trained independently to alleviate pressures of networks, thus they generate more realistic results.

\textbf{Few-shot face visually dubbing}. Few-shot face visually dubbing focus on repairing mouth region in source face according to driving audio. Existing works \cite{kr2019towards,prajwal2020lip,xie2021towards,park2022synctalkface,liu2022parallel} use a similar face decoder, multiple up-sample layers, to directly generate pixels of mouth region in source face. To improve visual quality, \cite{xie2021towards} utilize facial landmarks as intermediate structural representations. \cite{liu2022parallel} use effective SSIM loss. To improve lip-synchronization, \cite{prajwal2020lip} add one sync loss supervised by a pre-trained syncnet. \cite{park2022synctalkface} use one audio-lip memory to accurately retrieve synchronic lip shape. However, these direct-generation based methods fail to realize face visually dubbing on high resolution videos. Our method replaces direct-generation fashion with deformation to achieve more realistic results. 

\textbf{Person-specific talking face}. Person-specific talking face requires identity appears in training data. Early work \cite{suwajanakorn2017synthesizing} needs $16$ hours of footage to learn audio-lip mapping. \cite{fried2019text} reduce the training data to less than $1$ hour by rule-based selection. \cite{song2022everybody,thies2020neural,guo2021ad} utilize shared animation generators or Nerf \cite{mildenhall2020nerf} to make their methods only require several minutes footage. \cite{lahiri2021lipsync3d} use lighting normalization to handle the extreme condition of changeable light. \cite{ji2021audio,zhang2021facial} pay attention to realistic facial expressions, e.g., emotion and eye blink.

\begin{figure*}
\centering
\includegraphics[width=0.99\textwidth]{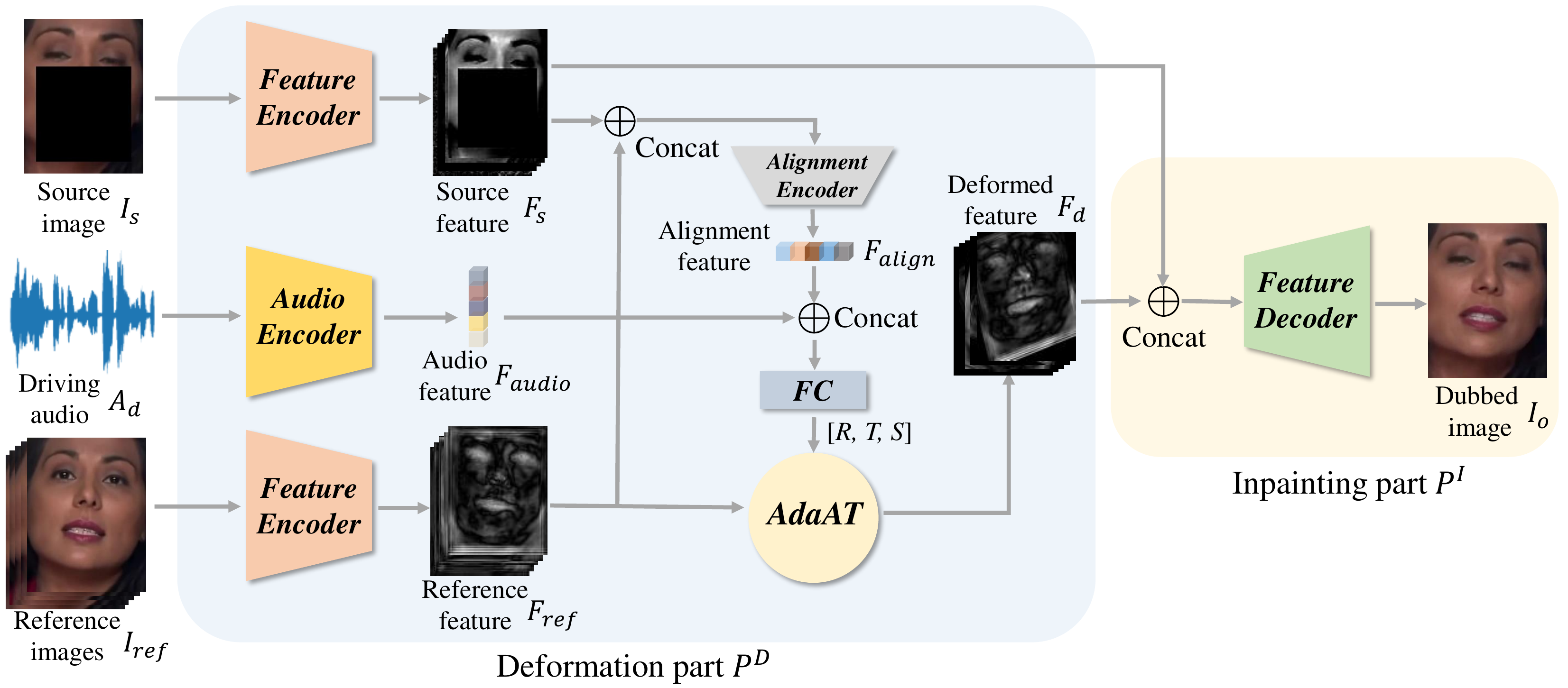}
\caption{Illustration of our DINet framework. DINet consists of one deformation part $P^{D}$ and one inpainting part $P^{I}$. In $P^{D}$, the feature maps of reference images are deformed spatially to synchronize mouth shape with the driving audio and align head pose with the source image. In $P^{I}$, deformed feature maps are used to inpaint the mouth region in the source face.}
\label{fig:pipeline}
\end{figure*}

\subsection{Spatial Deformation}
In deep learning based works, there are two main ways to realize spatial deformation: affine transformation and dense flow. In affine transformation based works, they first compute coefficients of affine transformations, then they do deformation on image feature maps by combining all affine transformations. \cite{jaderberg2015spatial,lin2017inverse} first import affine transformation in CNN networks. They compute one global affine transformation in each feature layer. To improve the complexity of deformation, \cite{siarohin2019first,wang2021one} increase the number of affine transformations by computing different coefficients in different 2D or 3D regions. \cite{zhang2022adaptive} propose one AdaAT operator to simulate a more complex deformation. They compute channel specific coefficients, increasing the number of affine transformations to hundreds. 

In dense flow based works, they directly use networks to compute a complete dense flow, then they warp feature maps with dense flow to achieve spatial deformation. The dense flow can be computed from graph convolutional networks \cite{yao2020mesh}, encoder-decoder networks \cite{ren2021pirenderer,doukas2021headgan} and  weight demodulation decoder \cite{wang2022latent}. In our work, we utilize AdaAT operator to realize spatial deformation because it synthesizes the best results. 

\section{Method}
We propose one DINet to achieve high resolution face visually dubbing. The structural details of DINet are shown in Figure \ref{fig:pipeline}. DINet consists of one deformation part $P^{D}$ and one inpainting part $P^{I}$. The deformation part $P^{D}$ focus on deforming the feature maps of the reference images spatially to synchronize the mouth shape with the driving audio and align the head pose with the source image. The inpainting part $P^{I}$ focus on utilizing the deformed results to repair the mouth region in the source face. We detail these two parts in the following.

\begin{figure*}
\centering
\includegraphics[width=0.98\textwidth]{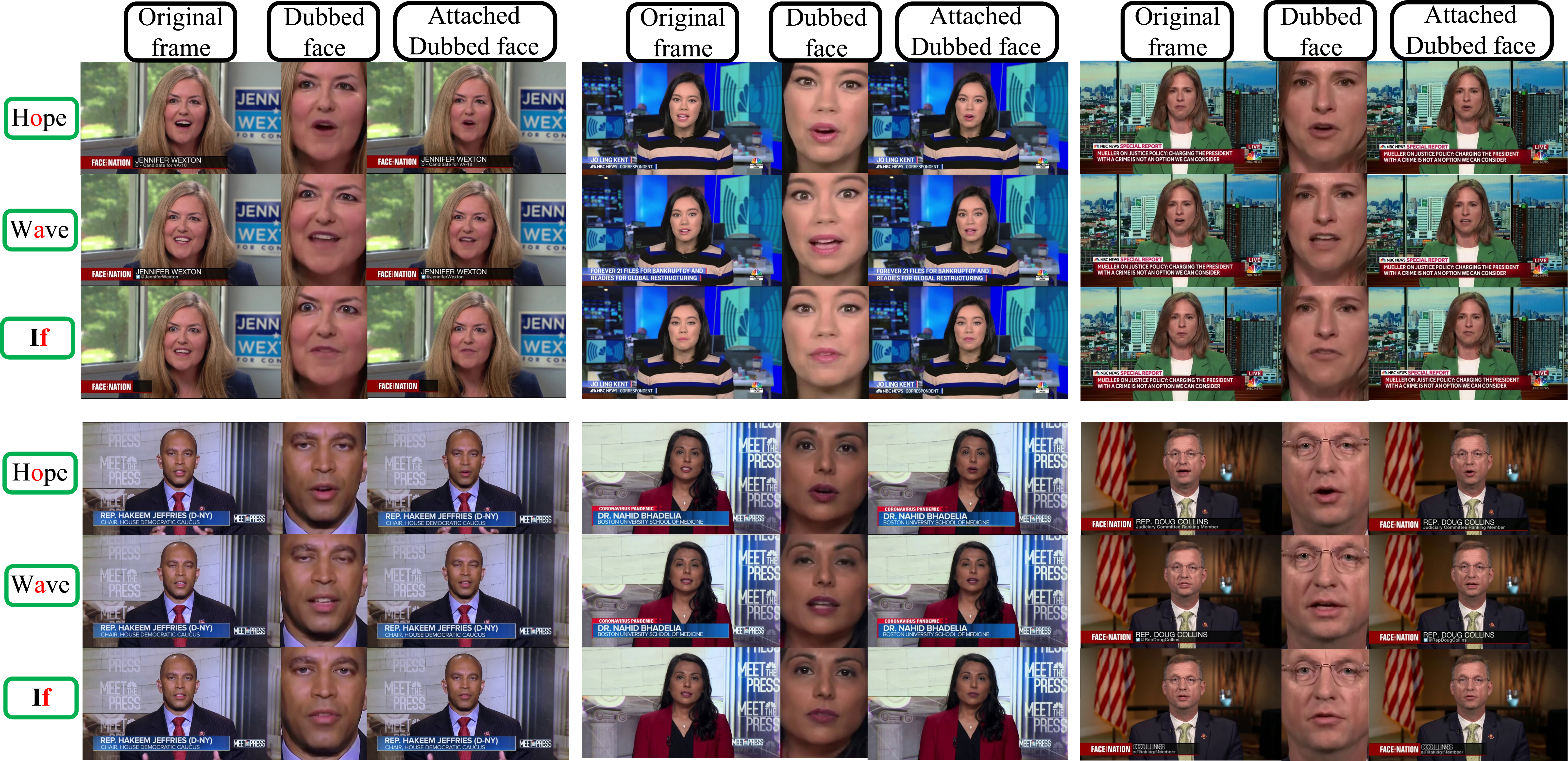}
\caption{The dubbed results of our DINet (Please zoom in for more details). Our method achieves realistic face visually dubbing on 1080P videos. More results can be observed in the demo video.}
\label{fig:synthetic results}
\end{figure*}

\subsection{Deformation Part}

The blue rectangle in Figure \ref{fig:pipeline} illustrates the structure of $P^{D}$. Given one source image $I_{s} \in R^{3 \times H \times W}$, one driving audio $A_{d} \in R^{T \times 29}$ ($29$ is a dimension of audio feature, we use deepspeech feature \cite{hannun2014deep} in our method) and five reference images $I_{ref} \in R^{15 \times H \times W} $, $P^{D}$ aims to produce deformed features $F_{d} \in R^{256 \times \frac{H}{4} \times \frac{W}{4}}$ that have synchronous mouth shape with $A_{d}$ and have aligned head pose with $I_{s}$. To realize this purpose, $A_{d}$ is first input into one audio encoder to extract audio feature $F_{audio} \in R^{128}$. $F_{audio}$ encodes the speech content of $A_{d}$. Then, $I_{s}$ and $I_{ref}$ are input into two different feature encoders to extract source feature $F_{s} \in R^{256 \times \frac{H}{4} \times \frac{W}{4}}$ and reference feature $F_{ref} \in R^{256 \times \frac{H}{4} \times \frac{W}{4}}$. Next, $F_{s}$ and $F_{ref}$ are concatenated and input into one alignment encoder to compute alignment feature $F_{align} \in R^{128}$. $F_{align}$ encodes the aligned information of head pose between $I_{s}$ and $I_{ref}$. Finally, $F_{audio}$ and $F_{align}$ are used to spatially deform $F_{ref}$ into $F_{d}$. 

In our method, we borrow the AdaAT operator \cite{zhang2022adaptive} instead of dense flow to realize spatial deformation. The main reason is that, compared with dense flow, AdaAT can deform feature maps with misaligned spatial layouts by doing feature channel specific deformations. AdaAT operator computes different affine coefficients in different feature channels. In our $P^{D}$, fully-connected layers are used to compute coefficients of rotation $R=\{\theta^{c}\}_{c=1}^{256}$, translation $T_{x}=\{t_{x}^{c}\}_{c=1}^{256}$/ $T_{y}=\{t_{y}^{c}\}_{c=1}^{256}$ and scale $S=\{s^{c}\}_{c=1}^{256}$. Then, these affine coefficients are used to do affine transformations on $F_{ref}$, as written in


\begin{equation} 
\begin{bmatrix} \hat{x}_{c} \\ \hat{y}_{c} 
\end{bmatrix}
=\begin{bmatrix} s^{c} cos(\theta^{c}) & s^{c}(-sin(\theta^{c})) & t_{x}^{c} \\ s^{c} sin(\theta^{c}) & s^{c} cos(\theta^{c}) & t_{y}^{c} \end{bmatrix}
\begin{bmatrix} x_{c} \\ y_{c} \\ 1 \end{bmatrix}, 
\label{eq:adaat}
\end{equation}
where $x_{c}$/$y_{c}$ and $\hat{x}_{c}$/$\hat{y}_{c}$ denote the pixel coordinates before and after affine transformation respectively. $c \in [1,256]$ represents $c_{th}$ channel in $F_{ref}$. After AdaAT operator, $F_{ref}$ is deformed into $F_{d}$.

\subsection{Inpainting Part}
The yellow rectangle in Figure \ref{fig:pipeline} illustrates the structure of $P^{I}$. $P^{I}$ aims to produce dubbed image $I_{o} \in R^{3 \times H \times W}$ from $F_{s}$ and $F_{d}$. To realize this purpose, $F_{s}$ and $F_{d}$ are first concatenated in feature channel. Then, one feature decoder with convolutional layers is used to inpaint the masked mouth and generate $I_{o}$. More structural details are in supplementary materials.

\subsection{Loss Function}
In training stage, we use three kind of loss functions to train DINet, including perception loss \cite{johnson2016perceptual}, GAN loss \cite{mao2017least} and lip-sync loss \cite{prajwal2020lip}.

\textbf{Perception loss.} Similar to \cite{zhang2022adaptive}, we compute perception loss in two image scale. Specifically, we downsample dubbed image $I_{o} \in R^{3 \times H \times W}$ and real image $I_{r} \in R^{3 \times H \times W}$ into $\hat{I}_{o} \in R^{3 \times \frac{H}{2} \times \frac{W}{2}}$ and $\hat{I}_{r} \in R^{3 \times \frac{H}{2} \times \frac{W}{2}}$. Then, paired images $\{I_{o},I_{r}\}$ and $\{\hat{I}_{o},\hat{I}_{r}\}$ are input into one pretrained VGG-19 network \cite{simonyan2014very} to compute perception loss. The perception loss is written as

\begin{small}
\begin{equation} \label{perceptual_loss}
\mathcal{L}_{p} = \sum_{i=1}^{N}\frac{{\|V_i(I_o)-V_i(I_r)\|}_1 + {\|V_i(\hat{I}_o)-V_i(\hat{I}_r)\|}_1}{2NW_iH_iC_i},
\end{equation}
\end{small}
where $V_{i}(.)$ represents $i_{th}$ layer in VGG-19 network. $W_iH_iC_i$ is the  feature size in $i_{th}$ layer.

\textbf{GAN loss.} We use effective LS-GAN loss \cite{mao2017least} in our method. The GAN loss is written as 
\begin{equation}
\label{gan_loss}
\mathcal{L}_{GAN}= \mathcal{L}_{D} + \mathcal{L}_{G}, 
\end{equation}
where
\begin{equation}
\mathcal{L}_{D} = \frac{1}{2}E(D(I_{r})-1)^{2} + \frac{1}{2}E(D(I_{o})-0)^{2}
\end{equation}
\begin{equation}
\mathcal{L}_G = E(D(I_{o}) - 1)^{2}.
\end{equation}
$G$ represents DINet and $D$ denotes discriminator. We use GAN loss on both single frame and five consecutive frames. The structural details of $D$ are in supplementary materials.

\textbf{Lip-sync loss.} As similar in \cite{prajwal2020lip}, we add a lip-sync loss to improve the synchronization of lip movements in dubbed videos. We replace audio spectrogram with deep speech feature and re-train the sycnet. The structural details of sycnet are in supplementary materials. The lip-sync loss is written as 
\begin{equation}
\mathcal{L}_{sync} = E(sycnet(A_{d},I_{o})-1)^{2}.
\end{equation}
We sum above losses as final loss $\mathcal{L}$, which is written as
\begin{equation}
\mathcal{L} = \lambda_{p}\mathcal{L}_{p} + \lambda_{sync}\mathcal{L}_{sync} + \mathcal{L}_{GAN}.
\end{equation}
where $\lambda_{p}$ and $\lambda_{sync}$ denote the weights of $\mathcal{L}_{p}$ and $\mathcal{L}_{sync}$. We set $\lambda_{p} = 10$ and $\lambda_{sync}=0.1$ in our experiment. 

\begin{table*}[]
\centering

{
\small
\centering
\begin{tabular}{lccccccccccc}
\hline
                    & \multicolumn{5}{c}{\textbf{HDTF}}          &  & \multicolumn{5}{c}{\textbf{MEAD-Neutral}}  \\ \cline{2-6} \cline{8-12} 
              & SSIM \(\uparrow\) & PSNR \(\uparrow\) & LPIPS\(\downarrow\) & LSE-D\(\downarrow\) & LSE-C\(\uparrow\) &  & SSIM\(\uparrow\) &PSNR\(\uparrow\) & LPIPS\(\downarrow\) & LSE-D\(\downarrow\) & LSE-C\(\uparrow\)  \\ \hline
ATVG        & 0.7315  & 20.5430 & 0.1104 & 8.4222 & 7.0611 &  & 0.8325 & 23.9723 & 0.0869 & 8.8908 & 5.8337 \\
Wav2Lip$-96$      & 0.9078  & 29.2875 & 0.0576 & 6.8714 & 8.2908 &  & 0.8994 & 28.5387 & 0.0779 & \textbf{6.0218}  & \textbf{8.9587}  \\
Wav2Lip$-192$                           & 0.8487  & 27.6561 & 0.1208 & 8.0912 & 6.9509 &  & 0.8036 & 25.6863 & 0.1302 & 7.8426  & 6.8515      \\
Wav2Lip$-384$                           & 0.8963  & 28.4695 & 0.0760 & 11.5373 & 3.5162 &  & 0.8415 & 25.9716 & 0.1166 & 9.8241  & 3.8001 \\
MakeitTalk    & 0.5969  & 19.8602 & 0.1592 & 11.4913 & 3.0293 &  & 0.8283  & 25.2988 & 0.1037 & 11.4256  & 2.5128  \\
PC-AVS              & 0.6383  & 20.6301 & 0.1077 & \textbf{6.6137} & \textbf{8.8550} &  & 0.7754  & 23.6950 & 0.0960 & 6.5035 & 8.6240   \\ \hline
Grount Truth                            & 1.0000  & N/A     & 0.0000 & 6.4989 & 8.9931 &  & 1.0000 & N/A     & 0.0000 & 6.2937 & 9.6747      \\
\textbf{DINet~(Ours)}                   & \textbf{0.9425}  & \textbf{30.0082} & \textbf{0.0289} & 8.3771 & 6.8416 &  & \textbf{0.9204} & \textbf{29.1177}  & \textbf{0.0459} & 7.5157 & 7.2603 \\ \hline
\end{tabular}
}
\caption{\label{tab:vssota} Quantitative comparisons with the state-of-the-art methods on talking face generation.}
\end{table*}

\section{Experiment}

In this section, we first detail datasets and implementation details in our experiment. Then, we show synthetic results of our method. Next, we carry out qualitative and quantitative comparisons with other state-of-the-art works. Next, we conduct ablation studies. Finally, an online user study is conducted to furthermore validate our method. 

\subsection{Datasets}
 We conduct experiments on two common high-resolution talking face datasets: HDTF dataset \cite{zhang2021flow} and MEAD dataset\cite{kaisiyuan2020mead}. 

 \textbf{HDTF dataset.} HDTF dataset contains about 430 in-the-wild videos collected with 720P or 1080P resolution. We randomly select 20 videos for testing in our experiments. 

 \textbf{MEAD dataset.} MEAD dataset records around 40 hours emotional in-the-lab videos at 1080P resolution. In our work, we do not focus on emotional face visually dubbing, so we select a total of 1920 videos with neutral emotion and frontal view as \textbf{MEAD-Neutral} dataset. In MEAD-Neutral dataset, we randomly select 240 videos of 6 identities for testing. 

 \begin{figure}
\centering
\includegraphics[width=0.42\textwidth]{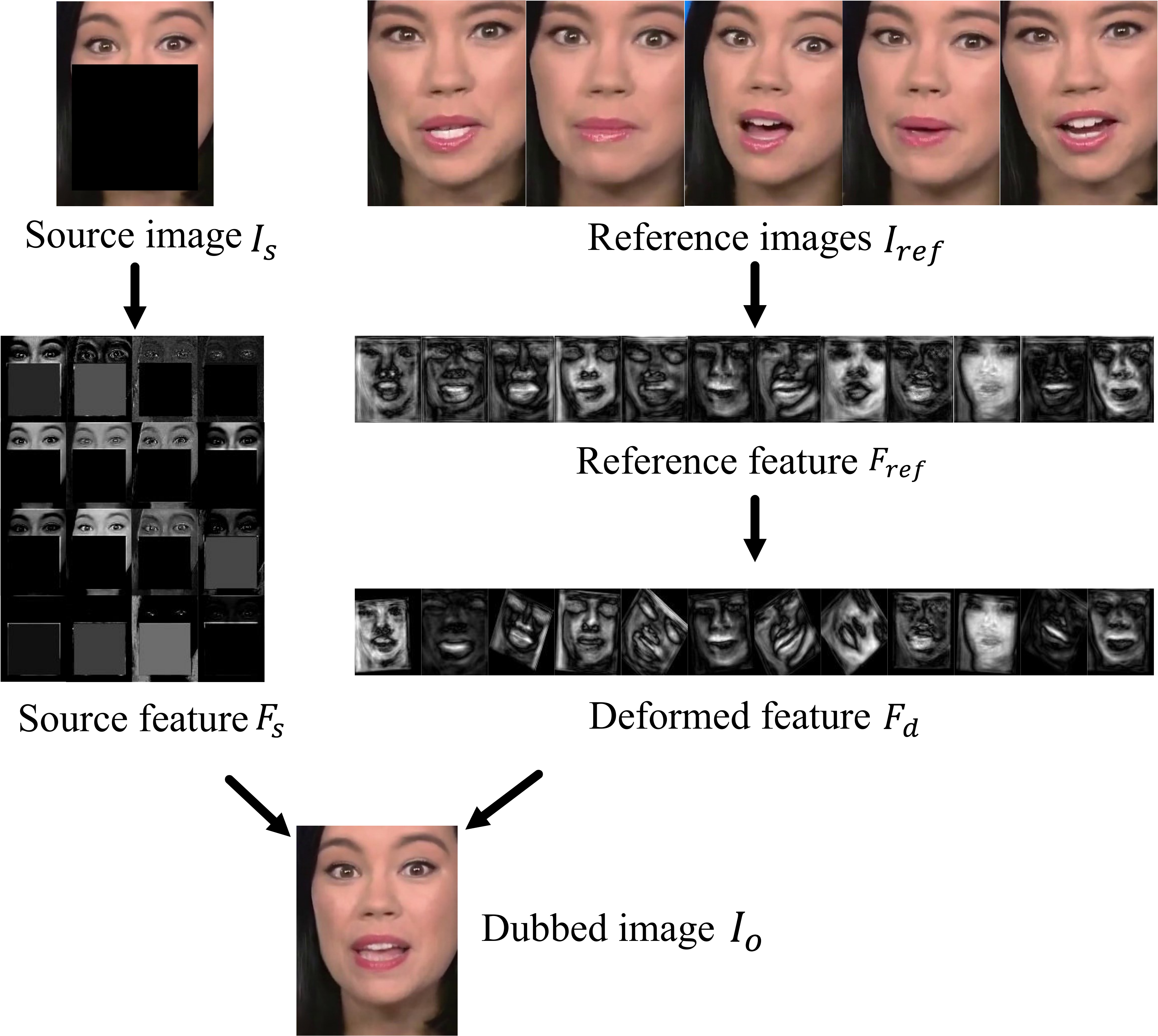}
\caption{Visualization of intermediate results in DINet, including the source image $I_{s}$, the reference images $I_{ref}$, the source feature maps $F_{s}$, the reference feature maps $F_{ref}$, the deformed feature maps $F_{d}$ and the dubbed image $I_{o}$.}
\label{fig:intermediate}
\end{figure}

\subsection{Implementation Details}
In data processing, all videos are resampled in 25 fps. We crop face region according to 68 facial landmarks of openface \cite{baltruvsaitis2016openface} and resize all faces into $416 \times 320$ resolution. The mouth region covers $256 \times 256$ resolution in resized facial image. More details about how we crop face are in supplementary materials. Considering that HDTF dataset and MEAD dataset have limited subjects, we use pre-trained deepspeech model \cite{hannun2014deep} to extract audio features of 29 dimention to improve generalization. The audio feature is aligned with video in 25 fps. 

In training stage, DINet inputs one source frame $I_{s} \in R^{3 \times 416 \times 320}$, one driving audio $A_{d} \in R^{5 \times 29}$ and five reference facial images $I_{ref} \in R^{15 \times 416 \times 320}$. Syncnet inputs $5$ frames of mouth images ($256 \times 256$) and corresponding deep speech features. We use Adam optimizer \cite{kingma2014adam} with default setting to optimize DINet and syncnet. The learning rate is set to $0.0001$. The batch size is set to 3 in DINet and 20 in syncnet on four A30 gpu. 


\subsection{Synthetic Results}

Figure \ref{fig:synthetic results} shows the dubbed results of our method. We display the original 1080P frames, the synthetic results of DINet and the original frames with attached synthetic face of six identities. Our method realizes realistic face visually dubbing on 1080P videos.   

Figure \ref{fig:intermediate} visualizes the intermediate results in DINet, including the source image $I_{s}$, the reference images $I_{ref}$, the source feature maps $F_{s}$, the reference feature maps $F_{ref}$, the deformed feature maps $F_{d}$ and the dubbed image $I_{o}$. It indicates that $F_{s}$ encodes spatial features both inside and outside the mask region. $F_{ref}$ encodes different spatial features of each frame in $I_{ref}$, so $F_{ref}$ is spatially misaligned. AdaAT performs different spatial deformation being specific to feature channels in $F_{ref}$ to generate $F_{d}$. The deformation is rich and contains scaling, rotation and translation, etc. 

\subsection{Comparisons with State-of-the-Art Works}
We compare our method with state-of-the-art one$-$/few$-$ shot talking head works that is open sourced, including ATVG \cite{chen2019hierarchical}, Wav2Lip \cite{prajwal2020lip}, MakeitTalk \cite{zhou2020makelttalk} and PC-AVS \cite{zhou2021pose}. Wav2Lip is the most relevant to our method, but their original model is trained on $96 \times 96$ resolution. For a fair comparison, we retrain their framework in $192 \times 192$ and $384\times 384$ resolution. We denote trained models in two resolutions as Wav2Lip-192 and Wav2Lip-384 respectively.  

\begin{figure*}[tb]
    \centering
    \includegraphics[width=0.97\textwidth]{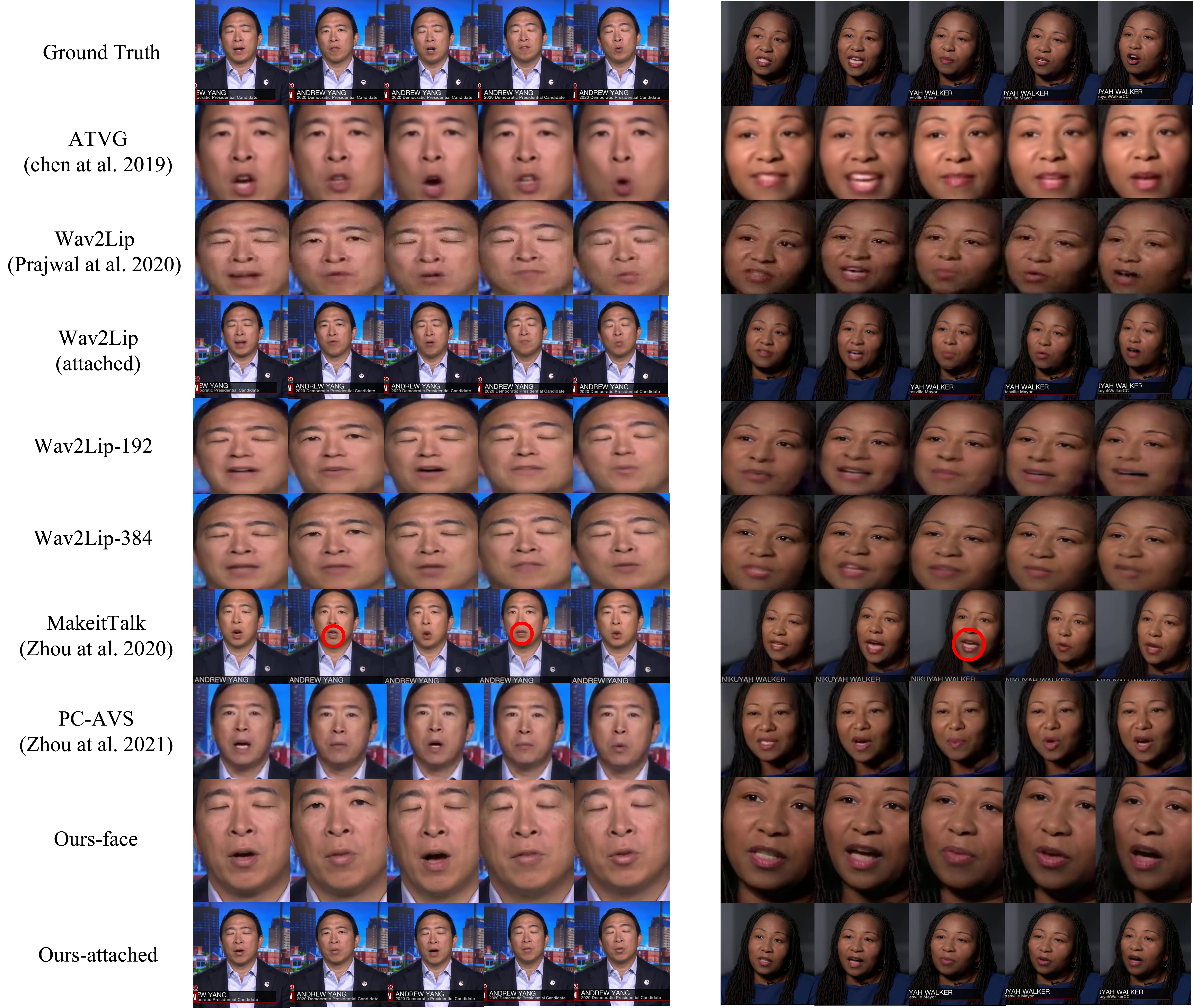}
    \caption{Qualitative comparisons with the state-of-the-art works~(Please zoom in for more details). Our method achieves high resolution face visually dubbing.}
    \label{fig:qua_comp}
\end{figure*}

\textbf{Qualitative comparisons.}
We first compare with state-of-the-art works in qualitative comparisons. Figure \ref{fig:qua_comp} illustrates the results. ATVG only generates $128 \times 128$ resolution videos, while our DINet can generate more realistic videos with $416 \times 320$ resolution. Wav2Lip synthesizes face with $96 \times 96$ resolution, and their mouth region becomes blurry when attaching the face into 1080P video. Wav2Lip-192 and Wav2Lip-384 still synthesize blurry results, although they are trained on high resolution videos.  The main reason is that wav2lip utilizes networks to dierctly generate pixels in mouth region, making networks easily neglect textural details. Our DINet deforms existing textural details into appropriate locations of mouth region, thus achieves more realistic results.

MakeitTalk tends to generate inaccurate mouth shape (see the red circle). This is caused by two main reasons:  their intermediate facial landmarks are not accurate enough, and facial landmarks are too sparse to describe lip motion details. In contrast, our DINet generates more accurate lip motions by directly learning a mapping between audio and mouth image. PC-AVS generates $224 \times 224$ resolution videos. Their resolution is limited due to direct-generation fashion. This limitation is statemented in their following work \cite{liang2022expressive}. Instead of directly generating pixels, our DINet utilizes operations of deformation and inpainting to realize high resolution face visually dubbing.

\textbf{Quantitative comparisons.}
Table~\ref{tab:vssota} shows results of quantitative comparisons. To evaluate the visual quality, we compute the metrics of Structural Similarity~(SSIM)~\cite{wang2004image}, Peak Signal to Noise Ratio~(PSNR), and Learned Perceptual Image Patch Similarity~(LPIPS)~\cite{zhang2018unreasonable}. To evaluate the audio-visual synchronization, inspired from \cite{prajwal2020lip}, we compute the metrics of Lip Sync Error Distance~(LSE-D) and Lip Sync Error Confidence~(LSE-C). In the evaluation of visual quality, our DINet gets the best results on all metrics. In the evaluation of audio-visual synchronization, our method is worse than Wav2Lip and PC-AVS. One possible reason is that metrical syncnet is trained on LRS2 dataset, while our model is trained from scratch.

\begin{figure}
\centering
\includegraphics[width=0.45\textwidth]{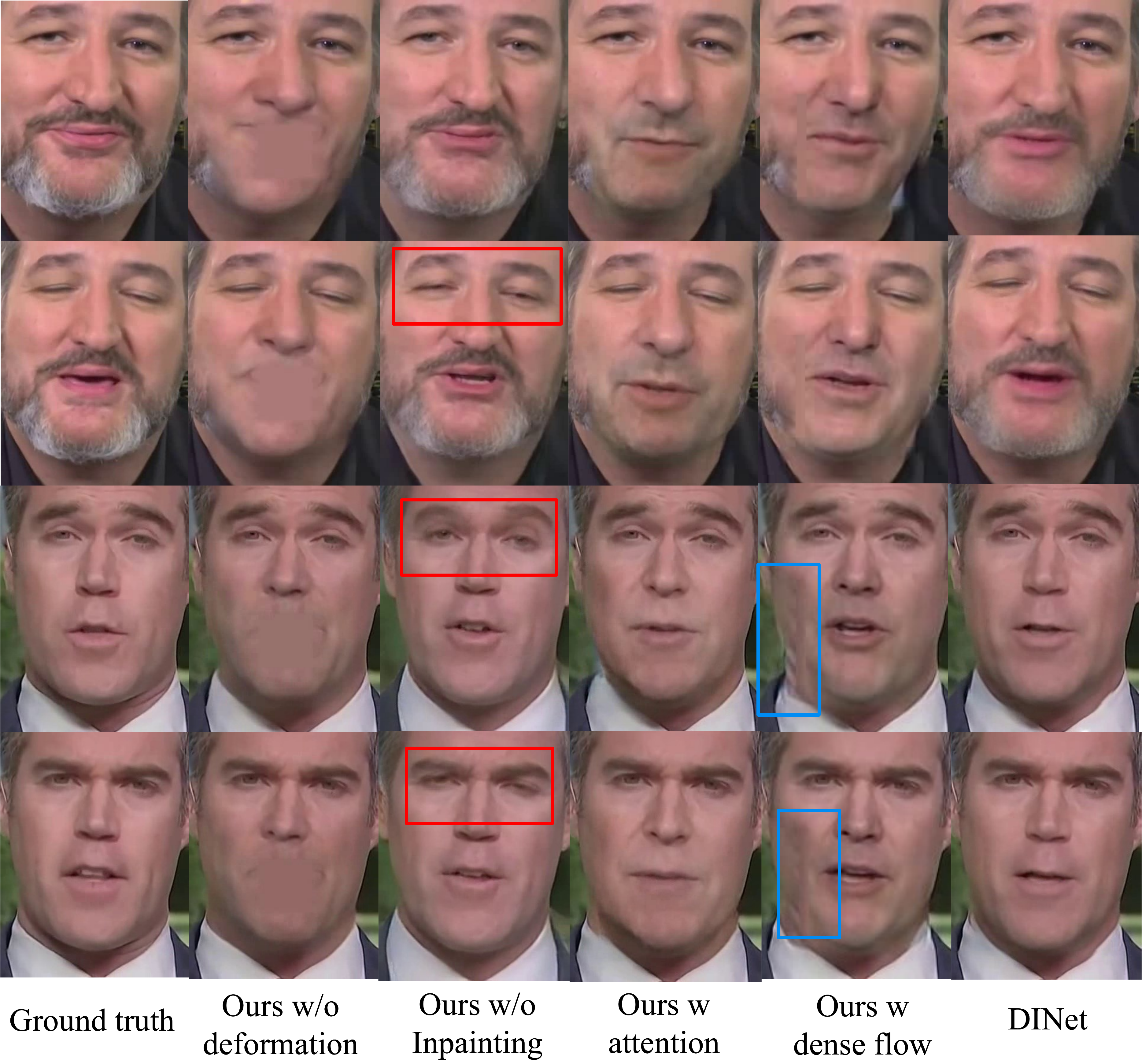}
\caption{Qualitative results of ablation study.}
\label{fig:ablation}
\end{figure}

\subsection{Ablation Study}
We conduct ablation experiments to validate each component in our DINet. Specifically, we set 5 conditions: (1) \emph{Ours w/o deformation}: we remove the AdaAT operation in DINet and inject $F_{audio}$ and $F_{align}$ into $F_{s}$ with AdaIN operation \cite{huang2017arbitrary}. (2) \emph{Ours w/o inpainting}: we directly generate $I_{o}$ from $I_{d}$ without fusing $F_{s}$. (3) \emph{Ours w attention}: we replace AdaAT operation with Attention \cite{vaswani2017attention} to select features from $F_{ref}$ instead of deformation. (4) \emph{Ours w dense flow}: we replace the AdaAT operation with dense flow to realize spatial deformation on $F_{ref}$. (5) \emph{Ours}: proposed DINet.

Figure \ref{fig:ablation} illustrates the qualitative results of ablation experiments. In the condition of \emph{Ours w/o deformation}, the synthetic facial images have blurry mouth. It indicates that it is difficult for networks to directly generate pixels in mouth region on high resolution videos. In the condition of \emph{Ours w/o inpainting}, the synthetic facial images have vivid textural details, including beard and nasolabial fold, due to effective deformation of AdaAT. However, without $F_{s}$ providing the information out of mouth region, upper face easily has incorrect performance, e.g., mismatched eyeblink and wrong sight line (see the red rectangle in Figure \ref{fig:ablation}) .

 In the condition of \emph{Ours w attention} and \emph{Ours w dense flow}, synthetic facial images lose more texture details. One possible reason is that "attention" and "dense flow" do "pixel-level deformation" while AdaAT operation do "region-level deformation". In "pixel-level deformation", each pixel is deformed flexibly into appropriate location and there is no regularization on the deformation. In "region-level deformation", region in a whole feature map do same affine transformation and the deformation is regularized by the number of feature channel. On one hand, too flexible deformation makes "attention" and "dense flow" are easily overfitting on the dataset. On the other hand, under a few-shot condition, mouth shape has higher correlations with driving audio than textural details, leading to networks focus on synthesizing synchronized lip motions instead of textural details. AdaAT has a strong regularization on the deformation, thus has a better performance on preserving textural details and facial structures than "attention" and "dense flow".

 Besides, \emph{Ours w dense flow} generate images with apparent artifacts in facial region (see blue rectangle in Figure \ref{fig:ablation}). One possible reason is that dense flow requires feature maps have similar spatial layouts. However, reference feature map $F_{ref}$ has different layouts due to the different head pose in reference images. Same deformation on misaligned feature maps may cause the generation of artifacts. 

 Table~\ref{tab:ablation} shows the quantitative results of ablation experiments. Our DINet gets the best visual quality according to the metrics of SSIM, PSNR and LPIPS.

 \subsection{User Study}
One user study is conducted for subjective evaluation. We randomly download five 1080P videos and one driving audio from internet. In one-shot works, we select the first frame as reference facial image and generate videos according to driving audio. In face visually dubbing works, we loop video frames if the length of video is shorter than driving audio. 20 volunteers are invited to rate realism of each synthetic video from 1~(pretty bad) to 5~(pretty good). The rating values are ATVG (2.9), Wav2Lip(2.4), MakeitTalk(3.1), PC-AVS (3.3) and ours (3.4). We get the best rating score.

\section{Limitations}
Our method achieves high-resolution face visually dubbing, yet we still suffer from several challenging conditions. Our DINet deforms reference images to inpaint mouth region in source face, so it can not handle conditions of changeable lighting, dynamic background, pendulous earrings, flowing hair and camera movements, and may generate artifacts out of face if mouth region covers background. The training videos in HDTF and MEAD-Neutral only have frontal view, so our method is restricted to limited head pose. Sometimes, the synthetic facial images are  sensitive to the selection of reference images. 

\begin{table}[]
\centering
\begin{tabular}{lcccc}
\hline
Method                  & SSIM\(\uparrow\)  &PSNR\(\uparrow\)   & LPIPS\(\downarrow\)   \\ \hline
Ours w/o deformation    & 0.9147        & 26.1665       & 0.0519            \\ 
Ours w/o inpainting     & 0.8691        & 26.0071       & 0.0561            \\ 
Ours w attention        & 0.9153        & 27.2451       & 0.0433            \\
Ours w dense flow       & 0.9001        & 23.5656       & 0.0656            \\ \hline
\textbf{Ours}           & \textbf{0.9425}        & \textbf{30.0082}       & \textbf{0.0289} \\ \hline
\end{tabular}
\caption{\label{tab:ablation} Quantitative results of ablation study on the HDTF dataset.}
\vspace{-0.5cm}
\end{table}

\section{Conclusion}
In this paper, we propose a Deformation Inpainting Network~(DINet), including one deformation part and one inpainting part, to realize high-fidelity face visually dubbing. The deformation part is designed to perform spatial deformation of feature maps of reference images to synchronize mouth shape with the driving audio and also align head pose with the source image. The inpainting part is designed to merge the deformed feature maps and the source feature map, to inpaint the mouth region in source image. With the combination of deformation and inpainting, our DINet preserves more textural details than the existing methods. Extensive qualitative and quantitative experiments have validated the performance of our method on high resolution face visually dubbing. In the future, we will make great efforts on solving above limitations.

\section{Acknowledgments}
This work is supported by the 2022 Hangzhou Key Science and Technology Innovation Program (No. 2022AIZD0054) and the Key Research and Development Program of Zhejiang Province (No. 2022C01011).

\bibliography{aaai23}

\end{document}